\documentclass[10pt,twocolumn,letterpaper]{article}
\usepackage{amsmath,amsfonts}

\usepackage{algorithm}
\usepackage{algorithmic}
\usepackage{amsmath, amssymb}
\usepackage{array}
\usepackage[caption=false,font=normalsize,labelfont=sf,textfont=sf]{subfig}
\usepackage{textcomp}
\usepackage{stfloats}
\usepackage{verbatim}
\usepackage{graphicx}
\usepackage{makecell}
\usepackage{multirow}
\usepackage{multicol}
\usepackage{booktabs}
\usepackage{color}
 \usepackage{cite}

\usepackage[final]{cvpr}      

\definecolor{cvprblue}{rgb}{0.21,0.49,0.74}
\usepackage[pagebackref,breaklinks,colorlinks,allcolors=cvprblue]{hyperref}

\def\paperID{1} 
\def\confName{CVPR Workshop MEIS}
\def\confYear{2025}

\title{SwarmDiff: Swarm Robotic Trajectory Planning in Cluttered Environments via Diffusion Transformer}

\author{
Kang Ding$^{1*}$ \quad
Chunxuan Jiao$^{1*}$ \quad
Yunze Hu$^{1}$ \quad
Kangjie Zhou$^{1}$ \quad
Pengying Wu$^{1}$ \quad
Yao Mu$^{2}$ \quad
Chang Liu$^{1\dagger}$ \\
$^{1}$Department of Advanced Manufacturing and Robotics, College of Engineering, Peking University, China \\
$^{2}$Department of Computer Science, The University of Hong Kong, Hong Kong SAR, China \\
\vspace{0.5em}
}

\begin{document}
\maketitle
{\renewcommand{\thefootnote}{}\footnotetext{\small * Equal contribution. $\dagger$ Corresponding author: \texttt{changliucoe@pku.edu.cn}.}}
\begin{abstract}
Swarm robotic trajectory planning faces challenges in computational efficiency, scalability, and safety, particularly in complex, obstacle-dense environments. To address these issues, we propose SwarmDiff, a hierarchical and scalable generative framework for swarm robots. We model the swarm’s macroscopic state using Probability Density Functions (PDFs) and leverage conditional diffusion models to generate risk-aware macroscopic trajectory distributions, which then guide the generation of individual robot trajectories at the microscopic level. To ensure a balance between the swarm's optimal transportation and risk awareness, we integrate Wasserstein metrics and Conditional Value at Risk (CVaR). Additionally, we introduce a Diffusion Transformer (DiT) to improve sampling efficiency and generation quality by capturing long-range dependencies. Extensive simulations and real-world experiments demonstrate that SwarmDiff outperforms existing methods in computational efficiency, trajectory validity, and scalability, making it a reliable solution for swarm robotic trajectory planning.
\end{abstract}    
\section{Introduction}
\label{sec:intro}

Accompanied by low-cost sensors, real-time communication, and computational advancements, swarm robotic systems have emerged as promising solutions for applications such as environmental exploration \cite{McGuire2019MinimalNS}, search and rescue \cite{dah2023search}, and coordination transportation \cite{9460560}. A key challenge is designing robust, flexible, and efficient trajectory planning strategies for large-scale swarms in obstacle-dense environments, where robots must navigate autonomously while maintaining global coordination under limited communication and computational constraints.



Existing trajectory planning methods for swarm robots fall into two main categories: microscopic and macroscopic approaches. Microscopic methods control each robot individually. While effective for small to medium-sized swarms, their computational cost becomes prohibitive for large-scale systems \cite{soria2021predictive, soria2021distributed, shome2020drrt, 10160164}. 
Macroscopic methods, in contrast, model the swarm as a whole to enhance scalability. For example, mean-field approaches \cite{9525253} represent swarm dynamics through averaged states, while density control methods \cite{9529040, 10111078} optimize swarm distributions via Monge-Kantorovich optimal transport. However, these methods often oversimplify motion models and lack obstacle avoidance mechanisms.
\begin{figure}[tbp]
    \centering
    \includegraphics[width=\linewidth]{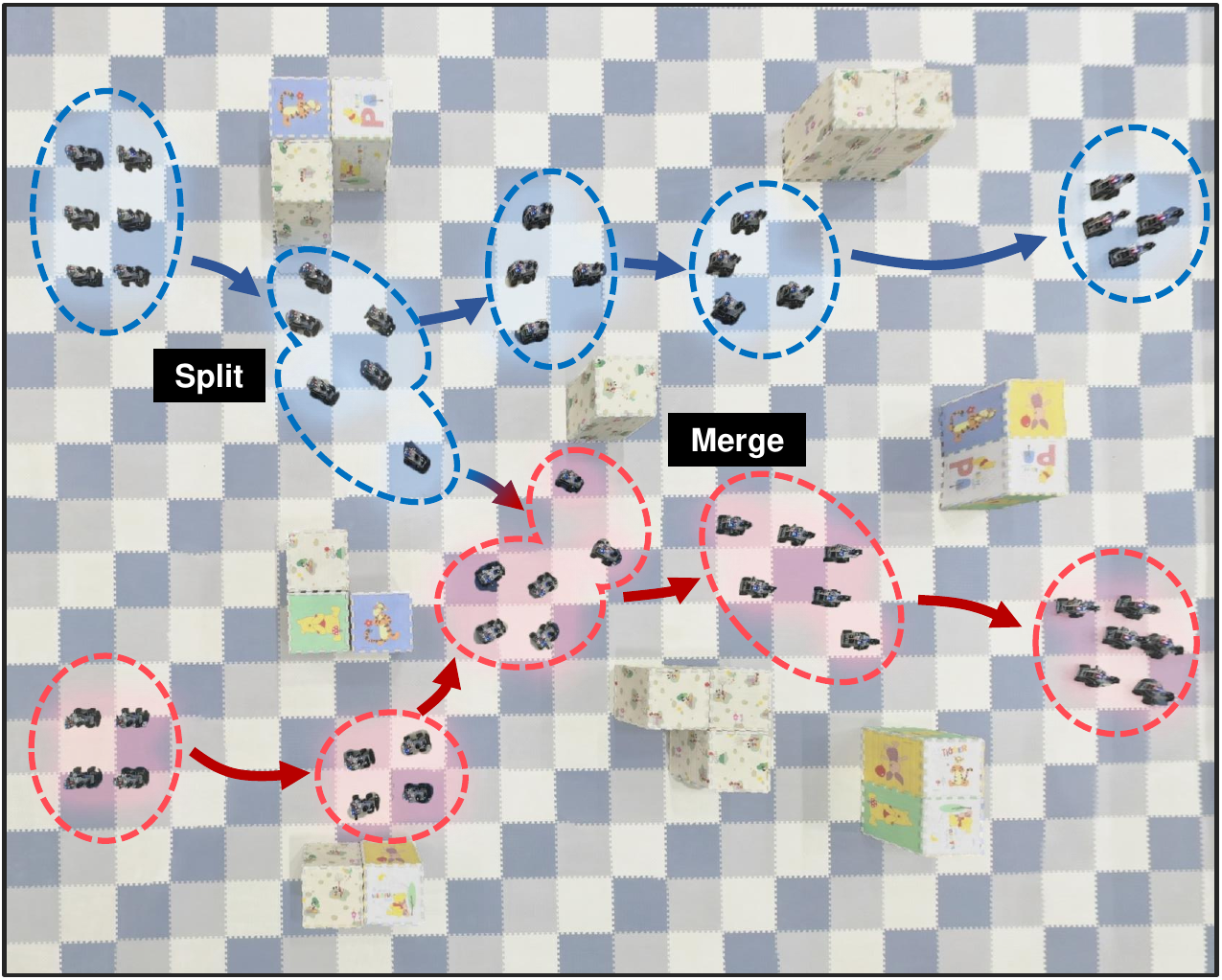}  
        \vspace{-15pt}
    \caption{Motion planning experiment with ten robots in a real-world obstacle-dense environment. The figure illustrates the robots' trajectories over five selected time steps, where each blue or red cluster represents the robot formations at a specific moment. The robots dynamically split and merge to navigate through the obstacles efficiently.}
    \label{fig:TEASER}
\end{figure}
To address the limitations of both approaches, hierarchical methods have gained attention as a promising alternative. These methods decompose the planning problem into multiple levels, where high-level strategies guide the overall swarm behavior, while low-level controllers handle individual robot motions. Multi-robot formation control \cite{alonso2017multi} optimizes global paths while refining local motion planning. However, these methods often impose rigid formation constraints, which may reduce adaptability in highly dynamic environments or when obstacles require flexible reconfiguration. The virtual tube approach \cite{mao2025tube, mao2024optimal} constrains swarm motion within predefined tubes ensuring smooth movement, but limiting flexibility in dynamic environments. 
ROVER \cite{rover} introduced a risk-aware planning approach that formulates a finite-time model predictive control problem based on the swarm’s macroscopic state, modeled as a Gaussian Mixture Model (GMM). However, its rigid probability-space discretization limits solution quality in dense environments and its optimization process is computationally expensive. SwarmPRM \cite{swarmprm} constructs probabilistic roadmaps in Gaussian space, ensuring asymptotic optimality for time-varying probability density functions and outperforming existing methods in path quality and efficiency. However, its reliance on state distribution sampling and connectivity checks reduces efficiency in complex environments, and its sampling constraints may prevent finding feasible paths in challenging scenarios.

Recent advances in generative models have demonstrated their capability to generate high-quality data across various domains \cite{karras2020analyzing, suzuki2022survey, chen2021decision}. In particular, generative sequence modeling \cite{chen2021decision, janner2021offline}, offers new solutions to planning problems by learning the joint distribution of state trajectories, control actions, rewards, and value functions in robotic systems. As a type of score matching-based generative model \cite{sohl2015deep, ho2020denoisingdiffusionprobabilisticmodels}, diffusion models have gained success in various domains with their strong conditional generation capabilities have explored robotic trajectory generation with diffusion models \cite{rasul2021autoregressive}. 
Diffuser \cite{janner2022planning} introduced a trajectory-level diffusion
probabilistic model enabling flexible constraints through reward guidance during sampling. Building on this, the works \cite{ajayconditional, ni2023metadiffuser, carvalho2023motion} employed diffusion models to generate decision trajectories conditioned on rewards or other auxiliary information, which are then used for planning. Additionally, \cite{wang2023diffusion} explored diffusion models for policy learning in reinforcement learning, while \cite{chi2023diffusion} focused on visuomotor control via action diffusion.
Most of the aforementioned methods are designed for single-robot scenarios. MMD \cite{shaoul2024multi} extended these approaches to multi-robot motion planning by integrating learned diffusion models with traditional search-based methods to generate data-driven motions while considering collision constraints. 
However, this method is currently limited to handling only dozens of robots, as the curse of dimensionality drastically expands the solution space, making trajectory optimization increasingly intractable with a growing number of robots. This computational burden renders the method unsuitable for large-scale robotic swarms.

To address these challenges, we propose SwarmDiff, a hierarchical, scalable, and computationally efficient generative framework for swarm robots. SwarmDiff leverages conditional diffusion models to learn risk-aware PDF distributions of macroscopic trajectories and design distributed motion control strategies for individual robots based on Model Predictive Control (MPC) (see Fig.~\ref{fig:TEASER}). The main contributions of this work are:
\begin{itemize}
    \item We propose SwarmDiff, enabling efficient and scalable motion generation for swarm robots. To the best of the authors' knowledge, this is the first approach that leverages generative diffusion models for large-scale swarm trajectory planning, providing a novel framework for learning and optimizing swarm motion.
   \item We design a cost gradient guidance mechanism that leverages the structure of swarm-distributed GMM trajectories, incorporating Wasserstein distance, Conditional Value at Risk (CVaR) and Gaussian process prior to steer the diffusion sampling process toward generating risk-aware, temporally coherent and task-adaptive GMM trajectories, enabling the model to flexibly adapt to diverse planning objectives.
    \item To address the limitation of existing temporal U-Net for trajectory generation tasks in capturing long-range dependencies, we introduce Diffusion Transformers (DiT) as denoising net, leveraging self-attention to model global structures and incorporating Wasserstein loss to improve trajectory coherence.
    \item Extensive simulations demonstrate that SwarmDiff outperforms baselines in computational efficiency and validity while providing strong scalability for swarm trajectory planning, the real-world experiments validate its feasibility and safety in practical scenarios.
\end{itemize}
\begin{figure*}[htbp]
    \centering
    \includegraphics[width=\textwidth]{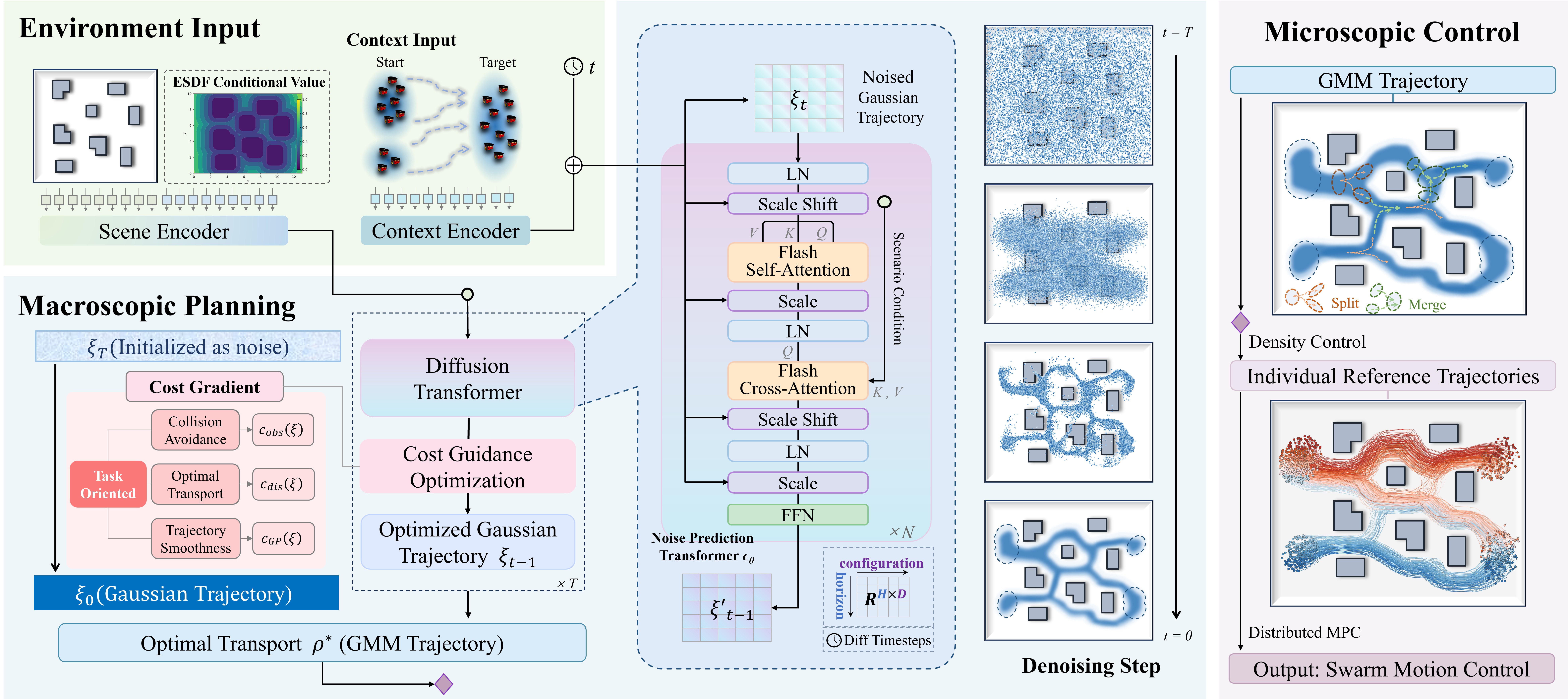}
   \caption{SwarmDiff overview. \textbf{Environment Input}: the Scene Encoder extracts obstacles map features and pre-calculated Euclidean Signed Distance Field (ESDF) as conditional values, and the Context Encoder processes the start and target distributions to provide contextual information. 
   \textbf{Macroscopic Planning}: In the bottom left (center), a  Diffusion Transformer iteratively refines an initial noisy Gaussian trajectory $\xi_T$ through a denoising process guided by cost gradients for task-oriented. This process results in an optimized Gaussian trajectory $\xi_0$, which is further refined via optimal transport to generate a GMM-based trajectory. The details of the DiT of SwarmDiff and the denoising process are illustrated in the middle. 
   \textbf{Microscopic Control}: individual reference trajectories are derived from the GMM trajectory by the Density Control method, and distributed MPC ensures coordinated swarm motion.}
    \label{fig:swarmdiff}
    \vspace{-10pt}
\end{figure*}
\section{Problem Formulation and Background}
\label{gen_inst}
\subsection{Hierarchical Swarm Robotic Motion Planning}
We address swarm motion planning in a 2D workspace \( W \subset \mathbb{R}^2 \), containing \( N_r \) robots and \( N_o \) static obstacles \( O = \bigcup_{i=1}^{N_o} O_i \). 
Let the initial positions of the swarm robots be denoted as \(Q = \{ q_1, q_2, \dots, q_{N_r} \}\), where \( q_i \) represents the initial position of the \( i \)th robot. 
The goal is to navigate the swarm from an initial to a target area. Let \(\tau = \bigcup_{i=1}^{N_r} \tau_i\) denote the set of all robot trajectories. Due to the complexity of optimizing individual trajectories for large-scale swarms, we propose a two-stage hierarchical framework: a \textit{macroscopic planning stage} models the swarm’s global state as a probability density function (PDF), capturing the collective motion trends and risk-aware collision avoidance constraints. The subsequent \textit{microscopic control stage} generates and tracks individual trajectories \(\tau_i\) based on this probabilistic representation, ensuring consistency with the global swarm behavior.

At the macroscopic planning stage, the state of the swarm is represented by a time-varying PDF \( \chi(x, t) \in \mathcal{P}(W) \), where \( x \in W \) and \( \mathcal{P}(W) \) is the space of PDFs supported on \( W \). 
This PDF describes the density distribution of robots across the workspace at time \( t \). To model this, we adopt the GMM representation:
\(\chi(x, t) = \sum_{j=1}^{N_k} \omega_j^k g_j^k,\)
where \( N_k \) is the number of Gaussian components \(g_j^k\) at time \( k \), \( \omega_j^k \geq 0 \) are the weights satisfying \( \sum_{j=1}^{N_k} \omega_j^k = 1 \).
The initial robot GMM distribution \( \chi(x, T_0) = \sum_{i=1}^{N_1} \omega_i^{T_0} g_i^{T_0} \) is calculated by the Expectation-Maximization (EM) algorithm based on the initial robot positions.
And we formulated the problem as an \textit{optimal transport problem} to transition the swarm from an initial PDF \( \chi(x, T_0) \) to a target PDF \( \chi(x, T_f) = \sum_{j=1}^{N_2} \omega_j^{T_f} g_j^{T_f}\) while minimizing costs.

As described in \cite{swarmprm}, the optimal transport between two GMMs can be achieved through transport weight
allocation between their respective Gaussian components.
Therefore, we assume that the optimal transport from \( \chi(x, T_0) \) to \( \chi(x, T_f)\)  can be achieved by \(K\) Gaussian trajectories \(\Xi = \{\xi_1, \dots, \xi_K\}\). Let \(\xi_i[t]\)  represent the state of the \(i-th\) Gaussian trajectory at the time \(t \in [T_0, T_f].\)
Each Gaussian trajectory \(\xi_i\) is weighted by \(\alpha_i,\sum_{i=1}^K \alpha_i = 1\).
Thus, the optimization objective is to find the optimal trajectories \( \xi \) and corresponding weights \( \alpha \) and is formulated as: \(
\min_{} \sum_{i=1}^K \alpha_i C(\xi_i)
\), where  \(C(\cdot)\) defines the cost function along a Gaussian trajectory.
Note that multiple Gaussian trajectories may spatially overlap at the same time step; however, such overlap does not indicate physical collision, but rather reflects the probabilistic fusion of Gaussian components.
As a result, the number of distinct Gaussian components \(N_k\) at time \(t_k\) can be less than or equal to the number of trajectories \(K\), i.e., \(N_k \leq K\).

At the microscopic control stage, the robots are assumed to be homogeneous, and their states are fully observable. The motion dynamics of the robots is modeled by a stochastic differential equation. Individual robots compute control inputs to follow the macroscopic PDF trajectory \( \chi(x, t) \) and avoid collisions with obstacles and other robots.

\subsection{Diffusion Models in Motion Planning Strategy}



Trajectory planning aims to generate smooth, collision-free trajectories in complex environments. Traditional approaches often struggle with high-dimensional and nonlinear constraints, making learning-based generative approaches an attractive alternative..
Denoising Diffusion Probabilistic Models (DDPMs, \cite{ho2020denoisingdiffusionprobabilisticmodels}) provide a powerful framework for trajectory generation by modeling the underlying distribution of feasible paths. These models operate through a two-step process:  
(1) a \textit{forward diffusion process}, which gradually adds noise to structured trajectories, and 
(2) a \textit{reverse denoising process}, which iteratively refines noisy trajectories into optimal trajectory.
The reverse process is learned by predicting the noise component at each step, minimizing the error \(\mathbb{E}_{t, x_0, \epsilon} \left[ \left\| \epsilon - \epsilon_{\theta}(x_t, t) \right\|^2 \right],\)
where \( x_t \) is a noisy version of \( x_0 \), following \( q(x_t | x_0) = \mathcal{N} \left( x_t; \sqrt{1 - \beta_t} x_0, \beta_t I \right). \)

To apply diffusion models to motion planning, we represent each trajectory as a sequence of Gaussian-distributed states \(\xi = \{\xi[t]\}_{t=0}^H\), where \(H\) is the time horizon, \(\xi[t] = [x_t, y_t, \sigma_{x_t}, \sigma_{y_t}, \rho_t]^\top\) denotes the position means \( \mu_t = (x_t, y_t) \) and the covariance matrices $\Sigma_t = \begin{bmatrix} \sigma_{x_t}^2 & \rho_t \sigma_{x_t} \sigma_{y_t},  
\rho_t \sigma_{x_t} \sigma_{y_t} & \sigma_{y_t}^2 \end{bmatrix}$.
We frame trajectory planning as a generative problem. Given an initial noisy trajectory, the goal is to iteratively refine it into an optimal, feasible trajectory \( \xi^* \) that minimizes a given cost function. 
By leveraging probabilistic inference, diffusion models efficiently generate feasible trajectories \( \xi_i \). The inference process follows a \( T \)-step denoising procedure, where an initial noisy trajectory \( \xi_T \) is gradually refined into an optimal one \( \xi_0 \).
\section{Swarm Robotic Motion Planner with Diffusion Transformer Model}
\label{headings}

SwarmDiff generates macroscopic path \(\rho^*\) for the optimal transport in GMM space that defines a probabilistic region and guides individual agents (see Fig.~\ref{fig:swarmdiff}). We depict the pseudo-code for training and two-stage trajectory generation for swarm motion planning via the diffusion transformer model as shown in \textit{Algorithm~\ref{alg:swarmdiff}}.
\vspace{-8pt}
\begin{algorithm}[H]
   \caption{SwarmDiff}
   \label{alg:swarmdiff}
\begin{algorithmic}[1]
   \STATE {\bfseries Input:} Start and goal GMMs, Initial swarm positions, Obstacles, Training dataset $\mathcal{D}$, Denoising function $\epsilon_\theta$
   \STATE {\bfseries Output:} Optimized swarm trajectories $\tau^*$
   
   \STATE {\bfseries 1. Training Phase:}
   \FOR{each training trajectory $\xi_j \in \mathcal{D}$}
      \STATE Add noise and train $\epsilon_\theta$ to predict denoising direction
   \ENDFOR

   \STATE {\bfseries 2. Macroscopic Trajectory Generation:}
   \FOR{each start component and goal component pair}
      \STATE Initialize noisy trajectories
      \FOR{reverse diffusion steps}
         \STATE Predict noise and update trajectory towards low-cost regions
      \ENDFOR
   \ENDFOR

   \STATE {\bfseries 3. Microscopic Trajectory Optimization:}
   \FOR{each robot}
      \STATE Extract individual trajectories from GMM 
      \STATE Apply MPC for trajectory tracking
   \ENDFOR

   \STATE {\bfseries Return:} Optimized swarm trajectories $\tau^*$
\end{algorithmic}
\end{algorithm}

\subsection{Diffusion Transformer for Macroscopic Path Generation}

\subsubsection{Inference for Generating Gaussian Trajectories}
\label{paragragh: inference}
The planning-as-inference paradigm reformulates motion planning as probabilistic inference \cite{janner2022planning, mukadam2018continuous}, where the posterior trajectory distribution \( p(\xi | \mathcal{O}) \) encodes both task objectives and prior feasibility constraints.  

Given a task objective \(\mathcal{O}\), motion planning diffusion models sample from the posterior:
\begin{equation}
\xi^* = \arg\max_{\xi} \log p(\xi | \mathcal{O}).
\end{equation}

Applying Bayes' rule, the posterior can be expressed as \(p(\xi | \mathcal{O}) \propto p(\mathcal{O} | \xi) p(\xi)\). Assuming that the likelihood factorizes into independent components, we have \(p(\mathcal{O} | \xi) \propto \prod_i p(o_i | \xi)^{\lambda_i}.\)
To ensure a well-defined probabilistic formulation, we assume all cost functions \( c_i(\xi) \) are non-negative and continuously differentiable. Under this assumption, each likelihood term can be modeled as an exponential function of the corresponding cost \(p(o_i | \xi) \propto \exp(-c_i(\xi))\).
Thus, we can rewrite the posterior distribution as \(p(\xi | \mathcal{O}) \propto \exp\left(-\sum_i \lambda_i c_i(\xi)\right) p(\xi).\)
Maximizing the posterior is equivalent to minimizing the negative log probability:
\begin{align}
    \xi^* &= \arg\min_{\xi} \left( -\log p(\mathcal{O} | \xi) - \log p(\xi) \right) \notag \\
          &= \arg\min_{\xi} \sum_i \lambda_i c_i(\xi) - \log p(\xi),
\label{eq:optimization}
\end{align}
Here, \(C(\xi) = \sum_i \lambda_i c_i(\xi)\) represents the weighted sum of individual cost terms, \( p(\xi) \) serves as a prior ensuring trajectory feasibility, and \( \lambda_\text{prior} \) balances prior adherence with task-oriented optimization.

Rather than relying on manually designed priors, we learn a data-driven trajectory prior \( p(\xi) \)  using a denoising diffusion model trained on a dataset of feasible trajectories. During inference, instead of directly optimizing for the Maximum a Posteriori (MAP) trajectory, we generate posterior samples by guiding the diffusion process with cost functions. This specific guidance is detailed in Sec.~\ref{subsec:sampling}.

\subsubsection{Integrating Transformers with Diffusion Models}
We introduce the DiT \cite{peebles2023scalable} as the denoising network in the reverse diffusion process, implicitly encoding a structural prior over the trajectory distribution through its self-attention mechanism.
Compared to existing diffusion models for trajectory generation that rely on temporal U-Net constrained by local receptive fields \cite{carvalho2023motion}, DiT utilizes self-attention to capture long-range dependencies and global trajectory structures. This is particularly beneficial for modeling high-dimensional motion patterns and long-horizon planning tasks.
Specifically, the model is formulated as $\epsilon_\theta$, which incorporates the initial and target distributions as context $\mathbf{c}$ and is trained to predict the noise $\epsilon$ introduced during the forward diffusion process.
The denoising process is parameterized by the mean function:
\(
\mu_\theta(\xi_t, t,\mathbf{c}) = \frac{1}{\sqrt{\alpha_t}} \left( \xi_t - \frac{1 - \alpha_t}{\sqrt{1 - \bar{\alpha}_t}} \epsilon_\theta(\xi_t, t, \mathbf{c}) \right)
\). This context guides trajectory generation, ensuring that the generated trajectories adhere to structural constraints while preserving diversity. The input consists of the noisy trajectory \( \xi_t \), the timestep \( t \), and context embeddings. 
To improve the quality of the trajectories, we adopt the Wasserstein distance as the loss function \(L(\theta) = \mathbb{E}_{i, \epsilon, \xi_0, \mathbf{c}} \left[ W(\epsilon, \epsilon_\theta(\xi_i, i, \mathbf{c})) \right]\), which offers a more structurally informative metric for trajectory similarity, as it assesses the global alignment of distributions. This leads to smoother and more coherent trajectories. 

Additionally, to enhance computational efficiency, we integrate Flash Attention \cite{dao2022flashattention}, which reduces the complexity of Transformer computations.

\subsubsection{Optimal GMM Trajectory Generation}


While the DiT model generates diverse and coherent Gaussian trajectories, these trajectories must be further refined to construct a globally optimal macroscopic representation. This necessitates formulating the trajectory aggregation process as a GMM optimization problem.

To achieve this, we employ the DiT model to generate multiple Gaussian trajectories, transitioning from \( N_1 \) initial Gaussian components to \( N_2 \) target components. For each initial-target pair, an optimal Gaussian trajectory is sampled.
Subsequently, we optimize the weights of these Gaussian trajectories to construct an optimal macroscopic GMM trajectory \( \rho^* \). This optimization is formulated as a Linear Programming (LP) problem:
\begin{equation}
    \Psi^* = \arg \min_{\Psi} \sum_{i,j} \Psi(i, j) L(g_i^{T_0}, g_j^{T_f}),
    \label{eq:LP}
\end{equation}
where \(g_i^{T_0}\) and \(g_j^{T_f}\) denote the initial and target Gaussian components and \( L(\cdot, \cdot) \) is the transport cost (Sec.~\ref{subsec:cost}). The transport policy \( \Psi \) is a matrix satisfying the constraints: \( \sum_{i=1}^{N_1} \Psi(i, j) = \omega_j^{T_f}, \quad 
    \sum_{j=1}^{N_2} \Psi(i, j) = \omega_i^{T_0}.\)

\subsection{Cost-Gradient-Guided Sampling}
\label{subsec:sampling}

Similar to classifier guidance \cite{ho2020denoisingdiffusionprobabilisticmodels}, which steers the sampling process by incorporating classifier gradients to enforce task-specific constraints, we propose Cost Gradient Guidance for Optimal Transport, biasing the trajectories towards the cost-optimal likelihood.

Specifically,  SwarmDiff directly samples from the posterior distribution \( p(\xi | \mathcal{O}) \), which is equivalent to drawing samples from the prior \( p(\xi) \) while optimally adjusting the trajectory to minimize the associated cost.
In the final denoising step, the trajectory \(\xi\) is represented as \(\xi_0\). Given the Markovian reverse diffusion process, the posterior distribution follows:
\(p(\xi_0 |  \mathcal{O}) = p(\xi_N) \prod_{t=1}^{N} p(\xi_{t-1}  |\xi_t, \mathcal{O}),\)
where \(p(\xi_N)\) is standard Gaussian noise. Following \cite{ho2020denoisingdiffusionprobabilisticmodels}, sampling from the task-oriented posterior \(p(\xi_0 | \mathcal{O})\) is
equivalent to sampling from a Gaussian distribution with
mean and covariance:
\begin{equation}
\xi_{t-1} \sim \mathcal{N}(\mu_t + \Sigma_t \mathbf{g}, \Sigma_t),
\end{equation}
where \(\mu_t\) is the estimated denoised mean, \(\Sigma_t\) is the noise covariance, and \(g\) is the gradient guiding the sampling process, computed from the cost function as \eqref{eq:optimization}:
\begin{equation}
\begin{aligned}
    \mathbf{g} = \nabla_{\xi_{t-1}} \log p(\mathcal{O} \mid \xi_{t-1}) \Big|_{\xi_{t-1} = \mu_t} 
     = -\sum_i \lambda_i \nabla_{\xi} c_i(\xi) \Big|_{\xi = \mu_t}.
\end{aligned}
\label{eq:gradient}
\end{equation}

This \textbf{cost gradient guidance} incorporates multiple weighted cost components to evaluate and optimize the macroscopic trajectories. Specifically, the term \( \mathbf{g} \) in Equation \ref{eq:gradient} aggregates these cost components, each designed to capture an aspect of trajectory quality.







\subsubsection{Collision Cost}




We use CVaR~\cite{rover} to model collision costs, ensuring risk-aware obstacle avoidance by penalizing high-risk swarm trajectories. To expedite inference, we incorporate the ESDF and its gradients as conditional information inputs, accelerating risk assessment.
ESDF provides a continuous representation of the environment, where each point \( p \) is assigned a signed distance \( s(p, O) \) to the closest obstacle \( O \). The gradient \( \nabla s(p, O) \) gives the normal direction to the nearest obstacle boundary.

For a Gaussian state of swarm with mean \( \mu \) and covariance \( \Sigma \), we approximate the signed distance function (SDF) using a first-order Taylor expansion around \( \mu \), yielding \(s(p, O) \sim \mathcal{N}(s(\mu, O), n^\top \Sigma n),\)
where \( n = \nabla s(\mu, O) / {\|\nabla s(\mu, O)\|} \) is the normal vector at \( \mu \). 

To quantify the risk of collision, we define \(\eta = -s(\mathcal{N}(\mu, \Sigma), O) \sim \mathcal{N}(-s(\mu, O), n^\top \Sigma n).\)
This represents the negative signed distance, where larger values indicate a higher probability of collision.
To ensure safety, we use CVaR at risk level \( \alpha \), which captures the expected worst-case signed distance beyond the \( \alpha \)-quantile:
\begin{equation}
    \text{CVaR}_\alpha(\eta) = -s(\mu, O) + \frac{\phi(\Phi^{-1}(1 - \alpha))}{\alpha} \cdot (n^\top \Sigma n),
\end{equation}
Here, \( \phi(\cdot) \) and \( \Phi(\cdot) \) are the PDF and CDF of the standard normal distribution, respectively.
Thus, given a safety margin \( \epsilon \), the collision cost function is:
\begin{equation}
   c_{\text{obs}}(\xi) = \sum_{j=0}^{H-1} \max(0, \text{CVaR}_\alpha(\eta^j) - \epsilon).
\end{equation}

\subsubsection{Transport Cost}
\label{subsec:cost}


We design the transport cost based on the Wasserstein metric from Optimal Mass Transport (OMT) theory \cite{chen2018optimal} to quantify the displacement of the swarm's Gaussian trajectory from its initial to target distribution. By minimizing the Wasserstein distance along the generated trajectories, we achieve optimal transport while preserving swarm continuity and reducing associated costs, thereby indirectly optimizing each robot's path.

For two Gaussian distributions \( g_1=\mathcal{N}(\mu_1,\Sigma_1) \) and \( g_2=\mathcal{N}(\mu_2,\Sigma_2) \), the Wasserstein distance \( W_2 \) \cite{chen2018optimal} is:
\begin{equation}
\begin{aligned}
W_2(g_1, g_2) = & \bigg \{ \Vert \mu_1-\mu_2 \Vert^2 \\
& + tr\left[\Sigma_1+\Sigma_2-2(\Sigma_1^{\frac{1}{2}}\Sigma_2\Sigma_1^{\frac{1}{2}})^{\frac{1}{2}}\right] \bigg \}^{\frac{1}{2}},
\label{eqn:GaussianWasserstein}
\end{aligned}
\end{equation}
where \( \Vert\cdot\Vert \) denotes the Euclidean distance and \( tr(\cdot) \) the matrix trace.  

For a trajectory \( \xi \), the total transport cost is defined as the cumulative sum of the Wasserstein distances between consecutive Gaussian states along the trajectory \(\xi = \{\xi[t]\}_{t=0}^H\). The total transport cost is given by:  
\begin{equation}
c_{dis}(\xi) = \sum_{j=0}^{H-1} W_2(\xi[j], \xi[j+1]).
\end{equation}






\subsubsection{Gaussian Process Cost}


We aim to ensure that each Gaussian trajectory evolves smoothly by introducing the Gaussian Process Cost as an optimization objective, penalizing trajectory discontinuities. For each \(\xi[t] = [x_t, y_t, \sigma_{x_t}, \sigma_{y_t}, \rho_t]^\top\), to explicitly model the evolution of covariance parameters, we define an extended state vector \( s_t =
\begin{bmatrix}
    x_t, y_t, \dot{x}_t, \dot{y}_t, \sigma_{x_t}, \sigma_{y_t}, \rho_t, \dot{\sigma}_{x_t}, \dot{\sigma}_{y_t}, \dot{\rho}_t
\end{bmatrix}^{\top} \in \mathbb{R}^{10}.\)

Following \cite{mukadam2018continuous}, we define the prior trajectory distribution as a zero-mean GP:  
\(q_F(\xi) = \mathcal{N}(\xi; 0, K)\), with constant time discretization \(\Delta t\). The covariance matrix  
\(K = [K(i, j)]_{0 \leq i, j \leq H-1}, \quad K(i, j) \in \mathbb{R}^{10 \times 10},\) where \(K(i, j) \in \mathbb{R}^{d \times d}\), captures temporal correlations.  

To penalize high curvature or discontinuities, we adopt the GP cost from \cite{carvalho2023motion}: 
\begin{equation}
c_{GP}(\xi) = \frac{1}{2} \sum_{t=0}^{H} \| \Phi_{t,t+1} s_t - s_{t+1} \|_{Q_{t,t+1}^{-1}}^2,
\end{equation}
where \(\Phi_{t,t+1}\) models state transitions, and \(Q_{t,t+1}\) encodes uncertainty or noise in the state, reflecting the correlation and variability between each dimension.

\subsection{Motion Planning with Microscopic Control}

Given the macroscopic GMM trajectory \(\rho^*(t)\), we derive reference trajectories for individual robots via density control, ensuring consistency with the GMM movement. A distributed Model Predictive Control (MPC) is then used for trajectory tracking.

\textbf{Density Control.}  
At time step \(T_i\), the robot positions are given by \(\mathcal{P}_{T_i} = \{\boldsymbol{p}_{k,T_i} \mid k \in \mathcal{R} \}\), where \(\mathcal{R}\) is the set of all robots.
To transition the swarm while preserving a target Gaussian Mixture Model (GMM) distribution at \(T_{i+1}\), we compute a transport map \(\mathcal{T}(\boldsymbol{x})\), modeled as an affine transformation in Gaussian space \cite{delon2020wasserstein}.
Once target positions are determined, each robot is assigned to a specific target via an Integer Linear Programming problem that minimizes total squared travel distance:
\begin{equation}
    \min_{\mathcal{A}} \sum_{k\in\mathcal{R}}\Vert\boldsymbol{p}_{k,T_i}-\boldsymbol{p}_{\mathcal{A}(k),T_{i+1}}\Vert^2,
    \label{eq:ILP}
\end{equation}
Here, \(\mathcal{A}(k)\) is a bijective mapping ensuring each robot moves efficiently while maintaining the GMM distribution.

By solving this assignment problem at each step, we obtain the discrete-time trajectories \(\tau^* = \{ \boldsymbol{p}_{k,T_i} \}_{k \in \mathcal{R}, i = 0, \dots, H},\)
guiding the swarm optimally from the initial to the target distribution.

\textbf{Model Predictive Control.}  
Based on the reference optimal trajectories \(\tau^*\), we use distributed MPC to ensure that each robot follows its assigned path.

Inspired by \cite{cheng2017decentralized}, we integrate Optimal Reciprocal Collision Avoidance (ORCA) with MPC to compute collision-free velocities over \(N_{MPC}\) steps. Additionally, large convex polytopes in free space are generated via a fast iterative method \cite{deits2015computing} to ensure feasibility.
\section{Simulations and Experiments}

We evaluate SwarmDiff through simulations and real-world environments, assessing its ability to handle complex swarm motion planning.





\subsection{Baselines and Metrics}
\textbf{Baselines:} 
\begin{itemize}
\item \textbf{SwarmPRM} \cite{swarmprm}: A probabilistic roadmap-based method for swarm motion planning.
\item \textbf{dRRT*} \cite{shome2020drrt}: A discrete RRT* approach designed for multi-robot planning.
\item \textbf{FC} \cite{alonso2017multi}: A multi-robot planning method that maintains predefined formations while navigating.
\item \textbf{SwarmDiff$^{A}$}: A SwarmDiff variant without ESDF conditional information, used to demonstrate the impact of environmental conditioning on performance.
\item \textbf{SwarmDiff$^{B}$}: A variant of SwarmDiff that replaces the Transformer with a temporal U-Net while maintaining the same parameters, enabling a direct comparison of the Transformer's advantages over the temporal U-Net in terms of time efficiency and generation quality.
\end{itemize}

\textbf{Evaluation Metrics:} Computational Time (\( T_{\text{sol}} \)) - Divided into macro-level and micro-level components: (1) Macro-Time (\( T_{\text{macro}} \)) - The time required for macroscopic trajectory planning, (2) Micro-Time(\( T_{\text{micro}} \)) - The time spent on individual robot trajectory calculations.
Average Trajectory Length (\( \bar{D} \)) -  The mean distance traveled by all robots from start to goal. Minimum Distance to Obstacles (\( d_{\text{obs}} \)) - The minimum distance between robots and obstacles. Minimum Distance Between Robots (\( d_{\text{rob}} \)) - The minimum distance between any two robots in the swarm. 
For all metrics, we report the mean of the planning results of three random times trials per context .

\textbf{Dataset Generation and Training:} 
SwarmDiff is trained on a dataset of swarm Gaussian trajectories generated by extended PRM \cite{swarmprm}. Each trajectory comprises 256 nodes, created in environments with randomly placed obstacles and randomly distributed start and goal positions. Each node is represented by a five-dimensional vector encoding the mean and covariance of a Gaussian distribution.

All implementations were in Python 3. Macroscopic planning was parallelized in PyTorch on an RTX 4090 GPU, while microscopic control ran on an Intel i9-12900K CPU with 32GB RAM.
    \vspace{-10pt}
\renewcommand{\arraystretch}{0.1}
\begin{table*}[ht]
    \centering
    \caption{Performance vs. Swarm Size}
     \vspace{-10pt}
    \small 
        \resizebox{\linewidth}{!}{
    \begin{tabular}{cccccccccc}
        \toprule
        \textbf{Robots} & \textbf{Method} & \multicolumn{4}{c}{\textbf{Environment I with Dense Obstacles}} & \multicolumn{4}{c}{\textbf{Environment II with Narrow Passages}} \\
        \cmidrule(lr){3-6} \cmidrule(lr){7-10}
        & & \textbf{$T_{\text{sol}}$ ($T_{\text{macro}}$)[s] $\downarrow$} & \textbf{$\bar{D}$[m] $\downarrow$} & \textbf{$d_{\text{obs}}$[m]} & \textbf{$d_{\text{rob}}$[m]} & \textbf{$T_{\text{sol}}$ ($T_{\text{macro}}$)[s] $\downarrow$} & \textbf{$\bar{D}$[m] $\downarrow$} & \textbf{$d_{\text{obs}}$[m]} & \textbf{$d_{\text{rob}}$[m]} \\
        \midrule
        & SwarmDiff & \textbf{1.30 (0.96)} & 179.0 & 3.1 & 0.264 & \textbf{1.43 (1.06)} & \textbf{176.0} & 3.3 & 0.264 \\
        & SwarmDiff$^{A}$ & 3.93 (3.57) & \textbf{177.2} & 2.5 & 0.300 & 2.67 (2.28) & 176.1 & 3.2 & 0.264 \\
        {20} & SwarmDiff$^{B}$ & 1.55 (1.21) & 181.3 & 2.6 & 0.264 & 1.69 (1.30) & 188.1 & 2.3 & 0.649 \\
        & SwarmPRM & 1308.84 (1267.39) & 193.3 & 4.7 & 0.611 & 960.24 (927.13) & 190.4 & 3.1 & 0.453 \\
        & $^{\triangle}$FC & 183.49 (25.03) & 284.9 & 2.0 & 0.240 & 176.31 (13.01) & 254.2 & 2.0 & 0.338 \\
        & $^{\triangle}$dRRT* & 2089.13 (/) & 203.4 & 0.8 & 1.6 & 2766.97 (/) & 235.5 & 1.0 & 0.583 \\
        \midrule
        & SwarmDiff & \textbf{1.82 (0.97)} & \textbf{177.9} & 2.4 & 0.264 & \textbf{1.99 (1.09)} & \textbf{175.6} & 2.6 & 0.264 \\
        & SwarmDiff$^{A}$ & 4.16 (3.34) & 178.6 & 2.5 & 0.264 & 3.03 (2.18) & 176.5 & 2.7 & 0.898 \\
        {50} & SwarmDiff$^{B}$ & 2.08 (1.24) & 180.3 & 1.4 & 0.264 & 2.08 (1.24) & 180.3 & 1.4 & 0.264 \\
        & SwarmPRM & 1364.45 (1288.63) & 197.8 & 2.4 & 0.303 & 954.72 (896.76) & 188.8 & 3.2 & 0.326 \\
        & $^{\triangle}$FC & 292.05 (28.44) & 274.8 & 1.1 & 0.240 & 302.60 (16.71) & 284.2 & 0.3 & 0.240 \\
        & $^{\triangle}$dRRT* & 6827.7 (/) & 377.2 & -5.0 & 0.43 & 5156.25 (/) & 408.5 & 1.0 & 0.500 \\
        \midrule
        & SwarmDiff & \textbf{2.59 (0.95)} & \textbf{178.0} & 1.6 & 0.264 & \textbf{2.72 (1.03)} & 176.3 & 2.2 & 0.264 \\
        & SwarmDiff$^{A}$ & 5.01 (3.37) & 178.3 & 2.4 & 0.264 & 4.03 (2.33) & \textbf{176.0} & 2.5 & 0.264 \\
        {100} & SwarmDiff$^{B}$ & 3.16 (1.34) & 181.1 & 1.8 & 0.264 & 2.96 (1.26) & 186.5 & 1.8 & 0.264 \\
        & SwarmPRM & 1536.08 (1398.07) & 191.5 & 4.3 & 0.289 & 1005.14 (902.41) & 191.3 & 2.7 & 0.289 \\
        & $^{\triangle}$FC & 508.75 (52.14) & 278.1 & 0.7 & 0.240 & 405.61 (13.55) & 254.3 & 0.9 & 0.240 \\
        & $^{\triangle}$dRRT* & - & - & - & - & - & - & - & - \\
        \midrule
        & SwarmDiff & \textbf{11.44 (1.00)} & 175.3 & 1.0 & 0.264 & \textbf{11.75 (1.03)} & \textbf{175.8} & 2.0 & 0.264 \\
        & SwarmDiff$^{A}$ & 13.94 (3.31) & \textbf{175.1} & 0.9 & 0.264 & 13.28 (2.25) & 176.2 & 2.0 & 0.264 \\
        {500} & SwarmDiff$^{B}$ & 11.91 (1.28) & 177.9 & 1.1 & 0.264 & 11.68 (1.23) & 186.5 & 1.2 & 0.264 \\
        & SwarmPRM & 2008.08 (1307.09) & 190.6 & 2.6 & 0.273 & 1396.24 (883.67) & 190.7 & 1.6 & 0.277 \\
        & $^{\triangle}$FC & 2129.87 (65.43) & 284.7 & 0.4 & 0.239 & 1645.82 (9.75) & 259.7 & 0.4 & 0.240 \\
        & $^{\triangle}$dRRT* & - & - & - & - & - & - & - & - \\
        \bottomrule
    \end{tabular}
    }
    \label{tab:number}
       \vspace{-10pt}
\end{table*}

\begin{table}[t]
\centering
\caption{Performance vs. Obstacle Densities}
\vspace{-5pt}
\resizebox{\linewidth}{!}{  
\begin{tabular}{@{}ccccccc@{}}
\toprule
Environment & Method & \textbf{$T_{\text{sol}}$ ($T_{\text{macro}}$)[s] $\downarrow$} & \textbf{$\bar{D}$[m] $\downarrow$} & \textbf{$d_{\text{obs}}$[m]} & \textbf{$d_{\text{rob}}$[m]}\\ 
\midrule
& SwarmDiff   & \textbf{11.05 (0.98)} & 171.5 & 2.4 & 0.264 \\
& SwarmDiff$^{A}$        & 12.26 (1.75) & \textbf{171.0} & 1.2 & 0.264 \\
{sparse}& SwarmDiff$^{B}$  & 11.81 (1.26) & 172.3 & 1.9 & 0.264 \\
& SwarmPRM   & 1017.43 (690.58) & 187.7 & 3.4 & 0.271 \\
& $^{\triangle}$FC         & 2082.90 (9.74) & 256.4 & 5.2 & 0.231 \\
\midrule
& SwarmDiff   & 11.58(\textbf{1.00}) & \textbf{179.8} & 1.4 & 0.264 \\
& SwarmDiff$^{A}$        & 13.00 (2.53) & 180.2 & 1.0 & 0.264 \\
{middle}& SwarmDiff$^{B}$  & \textbf{11.57} (1.23) & 183.8 & 1.7 & 0.264 \\
& SwarmPRM   & 1694.36(1166.06) & 193.9 & 3.2 & 0.271 \\
& $^{\triangle}$FC         & 2107.88 (38.15) & 299.8 & 0.3 & 0.240 \\
\midrule
& SwarmDiff   & \textbf{11.44 (1.00)} & \textbf{175.1} & 1.012 & 0.264 \\
& SwarmDiff$^{A}$        & 13.94 (3.31) & 175.3 & 0.913 & 0.264 \\
{dense}& SwarmDiff$^{B}$  & 11.91 (1.28) & 177.9 & 1.144 & 0.264 \\
& SwarmPRM   & 2008.08 (1307.09) & 190.6 & 2.593 & 0.273 \\
& $^{\triangle}$FC         & 2129.87 (65.43) & 284.6 & 0.353 & 0.239 \\
\bottomrule
\end{tabular}
}
\label{tab: obstacle}
\vspace{-10pt}
\end{table}




\subsection{Results in Simulations}

We evaluate SwarmDiff on varying sizes swarm robots, consisting of \( N = \{20, 50, 100, 500\} \) robots with a radius of \( r = 0.2m \) in two kinds of simulation environments within a bounded workspace \( W = [0, 200m] \times [0, 160m] \): \textbf{Environment I:} Maps with varying densities of randomly generated convex polygonal obstacles.
\textbf{Environment II:} A map featuring obstacles and narrow passages.
Swarm robots need to transition from an initial distribution to a target distribution while avoiding obstacles. 

SwarmDiff (including its ablation variants) and SwarmPRM were implemented in Python, while FC and dRRT* were run in MATLAB to leverage the official code provided in their paper. Methods executed in MATLAB are denoted with a superscript triangle (e.g., $^{\triangle}$method).

\begin{figure}[htbp]
    \centering
    \includegraphics[width=\linewidth]{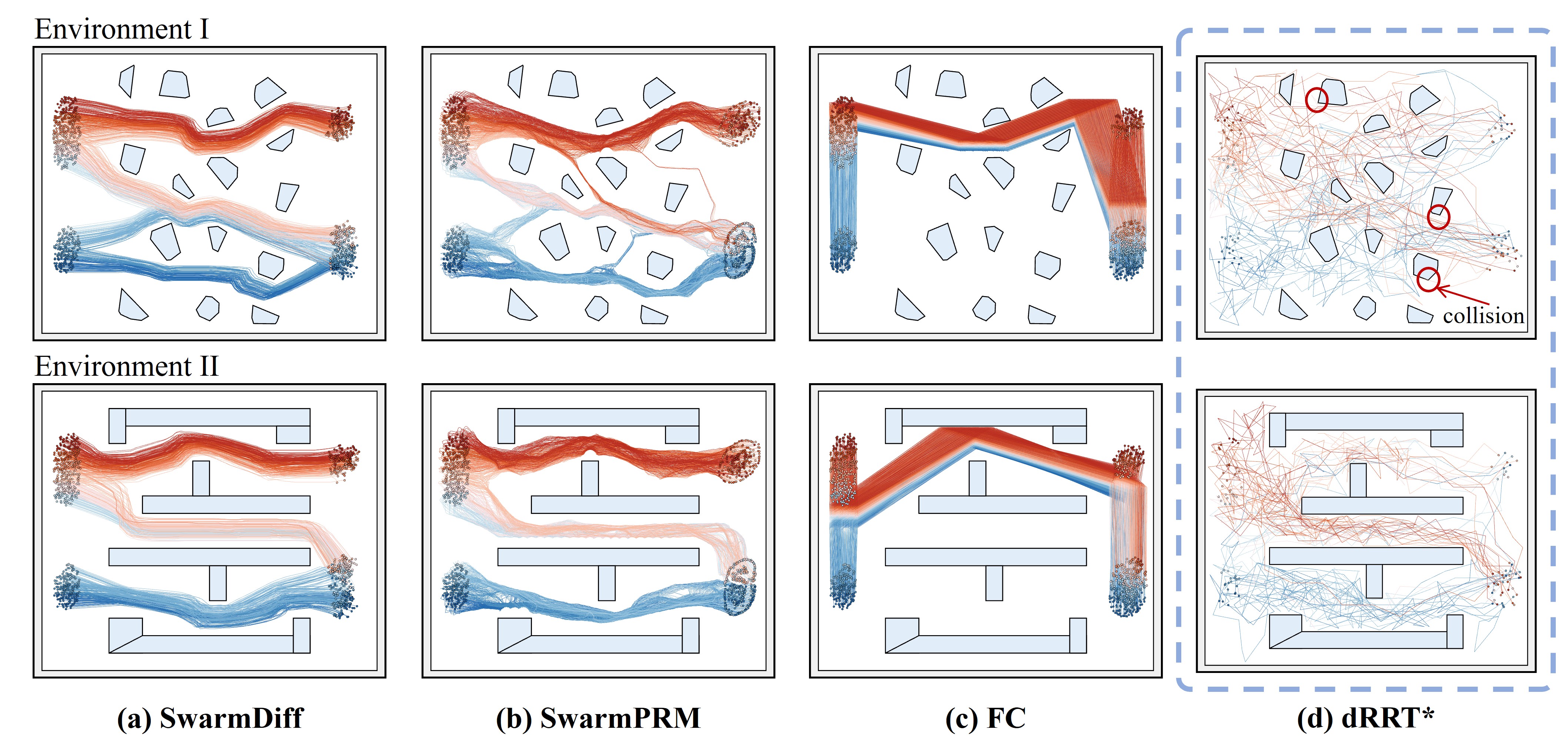}  
        \vspace{-20pt}
    \caption{\textbf{Trajectory comparison in environments I and II.}
    Trajectories of (a) SwarmDiff, (b) SwarmPRM, and (c) FC with $N = 500$ robots from the same initial positions, and (d) dRRT* with 50 robots sampled from the same distribution due to computational limits. Initial and final positions are colored circles, obstacles are black, and collisions (in dRRT*) are red.}
    \label{fig:sim_results}
    \vspace{-10pt}
\end{figure}

Our simulations are divided into two categories.
The first evaluates different approaches across various swarm sizes. Table~\ref{tab:number} summarizes the results. Compared to its ablation variants, the conditional version significantly improves computational efficiency. Additionally, our accelerated full version is both more efficient and expressive than the temporal U-Net with the same number of parameters.
Under identical conditions, SwarmDiff achieves significantly shorter computational times than SwarmPRM, demonstrating superior scalability. 
FC requires solving distributed nonlinear optimization problems at high frequencies to prevent collisions, causing computational time to grow rapidly with swarm size. dRRT* finds available solutions for up to 50 robots but fails for larger swarms in finite time due to exponential search space growth.
A comparative analysis across different swarm sizes underscores SwarmDiff’s capability to generate shorter and more optimized trajectories while maintaining computational efficiency. The trajectory planning results are presented in Fig.~\ref{fig:sim_results}.

The second set of experiments examines performance under varying obstacle densities in Table~\ref{tab: obstacle}. Statistical analysis shows that FC maintains a nearly constant path length, as all robots merge into a single formation before navigating obstacles. 
The computational time for the high-level planner to generate formation trajectories increases with the number of robots. However, as the planned trajectory guarantees formation safety, the computational cost of the low-level MPC remains largely unchanged, staying around 2000 seconds for a swarm of 500 robots.
Similarly, SwarmPRM exhibits little path length variation due to its ability to merge and split formations dynamically. However, as obstacle density increases, its computational cost rises significantly due to the computational cost of sampling-based probabilistic roadmap construction. Moreover, its Artificial Potential Field method used for local trajectory planning incurs additional computational cost, mainly due to the intensive calculations of attractive and repulsive forces and the high number of iterations required for convergence.
In contrast, SwarmDiff maintains stable path lengths and travel times across different obstacle densities, as its fixed denoising step makes it independent of environmental complexity. 
\vspace{-10pt}

\subsection{Results in Real-World Experiments}

\begin{figure}[h]
    \centering
    \includegraphics[width=\linewidth]{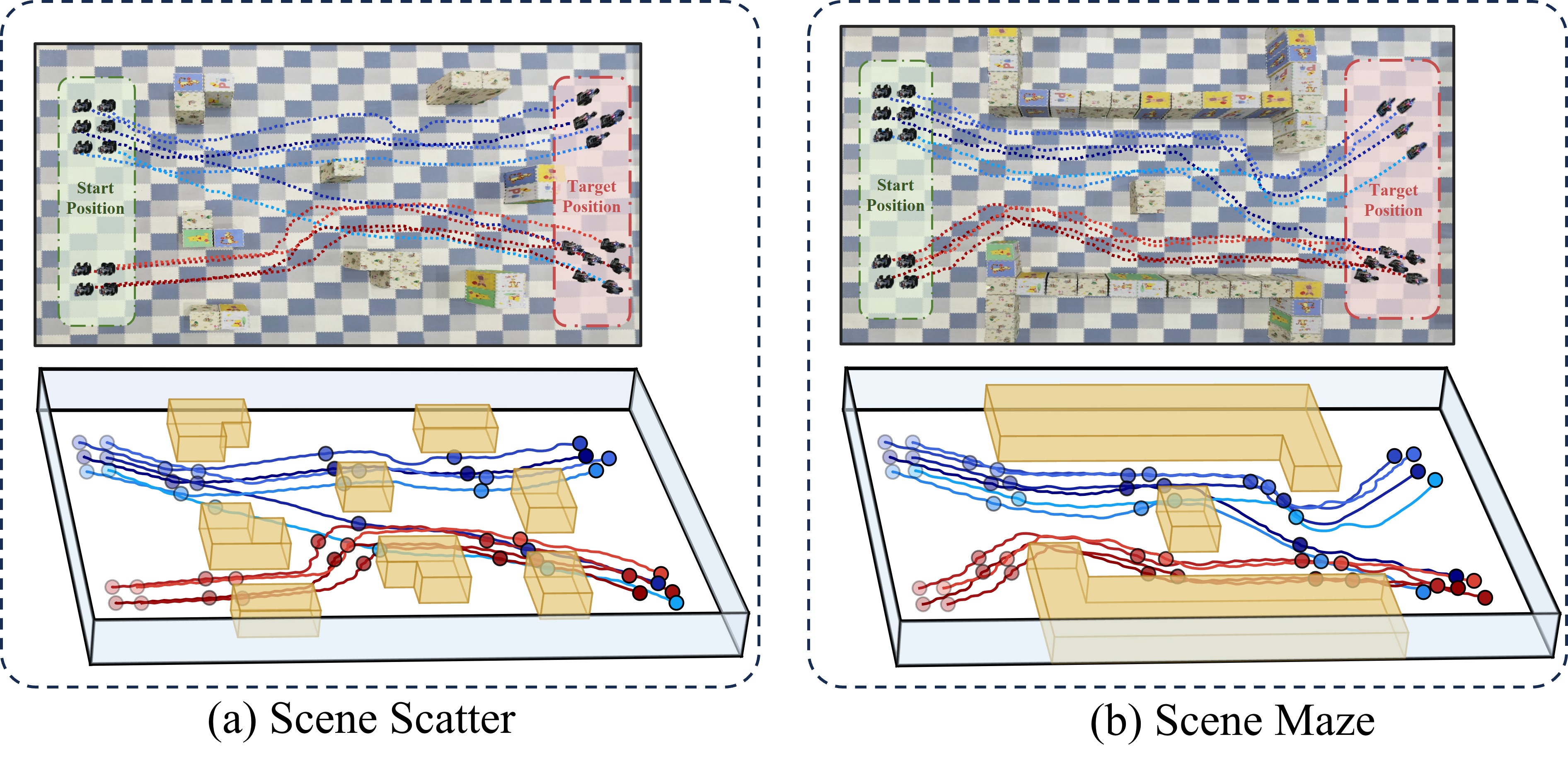}
        \vspace{-20pt}
    \caption{Experimental results of 10 robots in real-world environments. The top row shows real-world scenarios: Scene Scatter (left) and Scene Maze (right). The bottom row presents visualizations of real experimental data.}
    
    \label{fig:scene_comparison}
    \vspace{-5pt}
\end{figure}

To evaluate SwarmDiff in the real world, we conducted experiments using 10 Wheeltec R550 Ackermann robots with real-time localization provided by an FZMotion motion capture system. The robots followed SwarmDiff-generated trajectories, controlled via distributed MPC, while a ground station handled pose collection, trajectory planning, and ROS-based control.
Fig.~\ref{fig:scene_comparison} demonstrates that SwarmDiff consistently produces feasible trajectories. 



Table \ref{tab:real-world_metrics} presents key performance metrics from real-world experiments, demonstrating that the proposed method achieves both path efficiency and safety. The average path length $\bar{D}$ indicates that the method successfully navigates complex environments without unnecessary detours. Moreover, the minimum robot-obstacle distance $d_{\text{obs}}$ and the minimum inter-robot distance $d_{\text{rob}}$ are consistently above 0.25 m, ensuring safe motion and collision avoidance.
In addition to the previously mentioned metrics, we recorded the average computation time per MPC step ($T_{mpc}$) and the total task completion time ($T_{task}$), both serving as indirect indicators of the effectiveness of our high-level trajectory planning. Since the MPC controller follows the planned trajectories from SwarmDiff, the low per-step computation time and reasonable task duration suggest that the generated trajectories are well-optimized and feasible.
 \renewcommand{\arraystretch}{0.8}
\begin{table}[h]
    \vspace{-5pt}
    \centering
        \caption{Performance metrics in real-world experiments}
            \vspace{-5pt}
    \renewcommand{\arraystretch}{1.2} 
            \resizebox{\linewidth}{!}{
    \begin{tabular}{|l|c|c|c|c|c|}
        \hline
        \multirow{2}{*}{Scene} & \multirow{2}{*}{$\bar{D}$ [m]} & \multirow{2}{*}{$d_{\text{obs}}$ [m]} &  \multirow{2}{*}{$d_{\text{rob}}$ [m]} & \multirow{2}{*}{$T_{mpc}$ [s]} &\multirow{2}{*}{$T_{task}$ [s]} \\
        & & & & & \\
        \hline
        Scatter & 12.6423 & 0.2532& 0.3478  & 0.0094 & 173.9 \\
        \hline
        Maze & 13.6862 & 0.3291 & 0.3063  & 0.0083 & 182.0 \\
        \hline
    \end{tabular}
}
    \label{tab:real-world_metrics}
    \vspace{-10pt}
\end{table}
\section{Conclusion and Future Works}

This paper proposes SwarmDiff, a scalable and computationally efficient framework for trajectory generation of swarm robots. It models swarm macroscopic state as PDF and uses DiT to produce Gaussian trajectories. By incorporating Wasserstein metrics and CVaR, SwarmDiff enables risk-aware, optimal GMM planning, facilitating distributed motion control at the individual level.  Simulations and real-world tests show superior efficiency, scalability, and flexibility over existing methods. Future work will focus on real-time adaptive learning for dynamic environments.

{
\newpage
    \small
    \bibliographystyle{ieeenat_fullname}
    \bibliography{main}
}

\end{document}